\documentclass{article}

\usepackage[final,nonatbib]{nips_2018}
\usepackage[sorting=none, style=numeric-comp]{biblatex}
\bibliography{references}

\usepackage[utf8]{inputenc} 
\usepackage[T1]{fontenc}    
\usepackage{hyperref}       
\usepackage{url}            
\usepackage{booktabs}       
\usepackage{amsfonts}       
\usepackage{nicefrac}       
\usepackage{microtype}      
\usepackage{graphicx}
\usepackage{caption}
\usepackage{subcaption}
\usepackage[titletoc,title,page]{appendix}
\usepackage{float}
\usepackage[shortlabels]{enumitem}

\usepackage[colorinlistoftodos]{todonotes}

\title{Rethinking clinical prediction: Why machine learning must consider year of care and feature aggregation}

\author{
  Bret Nestor \\
  University of Toronto\\
  \texttt{bretnestor@cs.toronto.edu} \\
  \And
  Matthew B. A. McDermott\\
  MIT\\
  \texttt{mmd@mit.edu}\\
  \And
  Geeticka Chauhan\\
  MIT\\
  \texttt{geeticka@mit.edu}\\
  \AND
  Tristan Naumann\\
  Microsoft Research\\
  \texttt{tristan@microsoft.com}\\
  \And
  Michael C. Hughes \\
  Tufts University \\
  \texttt{mhughes@cs.tufts.edu} \\
  \And
  Anna Goldenberg\\
 Hospital for Sick Children, University of Toronto,\\ Vector Institute\\
  \texttt{anna.goldenberg@utoronto.ca}\\
  \And
  Marzyeh Ghassemi \\
  University of Toronto, Vector Institute\\
  \texttt{marzyeh@cs.toronto.edu} \\
}

\begin{document}

\maketitle

\begin{abstract}
Machine learning for healthcare often trains models on de-identified datasets with randomly-shifted calendar dates, ignoring the fact that data were generated under hospital operation practices that change over time.
These changing practices induce definitive changes in observed data which confound evaluations which do not account for dates and limit the generalisability of date-agnostic models.
In this work, we establish the magnitude of this problem on MIMIC, a public hospital dataset, and showcase a simple solution. We augment MIMIC with the year in which care was provided and show that a model trained using standard feature representations will significantly degrade in quality over time. We find a deterioration of 0.3 AUC when evaluating mortality prediction on data from 10 years later. We find a similar deterioration of 0.15 AUC for length-of-stay. In contrast, we demonstrate that clinically-oriented aggregates of raw features significantly mitigate future deterioration.
Our suggested aggregated representations, when retrained yearly, have prediction quality comparable to year-agnostic models.
\end{abstract}

\section{Introduction}
Training predictive machine learning systems for clinical outcomes has been made possible by the shift towards electronic health records (EHR) in modern healthcare systems. The best example of this is the widely-used MIMIC-III dataset~\cite{johnson2016mimic}, which is a public \emph{de-identified} dataset from a single hospital's intensive care units (ICU). 

A key part of de-identification is obscuring any calendar dates related to care, e.g., MIMIC-III dates are shifted into the future by a random offset for each individual patient, resulting in new dates ``between the years 2100 and 2200''~\cite{johnson2016mimic}. Not accounting for such shifts is problematic because EHR data is fundamentally generated as a byproduct of care. 
Care practices evolve over time via concept drift~\cite{concept_drift2010} and are serious in their own right~\cite{lazer2014parable}.
MIMIC notably reflects such a change, through the EHR system update from Carevue to Metavision in the Beth Israel Deaconess Medical Center in 2008 \cite{johnson2016mimic}. However, even without an EHR change, a \emph{random date shift} may also cause models to unintentionally \emph{train} with data generated from newer care practices than they are tested on. 

Previous work has trained models from MIMIC-III data to predict outcomes such as mortality~\cite{ghassemi2014unfolding,cheRecurrentNeuralNetworks2018}, length-of-stay~\cite{harutyunyan2017multitask} or billing codes~\cite{choi2017gram}. However, it is standard machine learning practices to sample a random set of patients for training and test splits, and select a large number of raw features as input. Our goal in this paper is two-fold: 1) to investigate whether date randomisation interferes with evaluation of predictive models trained with standard practices, and 2) to identify the extent that standard health tasks are meaningful over evolving care practices. 

We show that standard machine learning practices on the date-randomised data from MIMIC over-estimates predictive model performance on ICU mortality prediction - a common task. We further demonstrate that simple feature aggregation can mitigate degradation in predictive power over time. We identify that predictive performance saturates quickly in mortality prediction, and further demonstrate that other tasks show a similar performance pattern, and improvement to generalisation as a result of feature aggregation. 

\section{Data}
We use MIMIC III, a public dataset with EHR data from over $58,900$ hospital admissions of nearly $38,600$ adults at Beth Israel Deaconess Medical Center from 2001 to 2012 \cite{johnson2016mimic}. We consider the first intensive care unit (ICU) stay of patients older than 15 who were in the ICU for at least 36 hours ($21,877$ unique ICU stays). We extract patient demographic data, as well as physiological measurements queried from the \texttt{CHARTEVENTS} table, which contains the measurements captured at patient bedside. We use the same steps taken to pre-process data as found in \cite{suresh2017clinical,mcdermott2018semi}. In addition to the publicly-available MIMIC data, a Limited Data Use Agreement was granted to provide the admitting year for these patients.

We use the first 24 hours of data for each patient, and collect physiological measurements into hourly buckets via averaging. Several works have focused on imputation methods for healthcare data \cite{cheRecurrentNeuralNetworks2018,trespSolutionMissingData1997, yoonGAINMissingData2018, janssenMissingCovariateData2010a, agnielBiasesElectronicHealth2018, lin2008exploiting}. 
We use simple imputation to assign three sub-features to the data: the imputed (forward-filled) measurement of the feature, a binary indicator of whether or not that feature was observed at that time, and the number of hours since the feature was last observed\cite{cheRecurrentNeuralNetworks2018}. 

\section{Methods}
\paragraph{Data Representation}
For each patient, we investigate the impact of two common data representation strategies on tasks. 

\begin{enumerate}[nosep]
\item \texttt{Item-ID} representation:
First, we allow measurements of each variable's own raw \texttt{itemid}\footnote{The \texttt{itemid} is the internal MIMIC identifier for different lab measures, vitals, etc.}, which is a common strategy in prediction. 
\item \texttt{Clinically Aggregated} representation:
While prior work has sought to learn mappings between similar features\cite{Gong:2017dh, 2018arXiv180107860R}, we use simple aggregations of highly-present \texttt{itemid}s~\cite{harutyunyan2017multitask} to create ``expert driven'' condensed categories\footnote{An example of this aggregation is given in Appendix~\ref{app:feat_agg}}.
\end{enumerate}

\paragraph{Models}
We use a random forest (RF) classifier for all tasks with simple imputation to handle missing data, as described in \cite{cheRecurrentNeuralNetworks2018}. Depending on whether a), the test years overlap with the training years, or b), the training years completely precede the test years, then variation will be introduced by random train/test splits (80\% and 20\%) and random initialisations, respectively.

In all cases, 5-fold cross validation was applied to the training data, using a random search to find best parameters for maximum area under the receiver-operator curve (AUROC) on the validation split. It is important to note that since the variation is introduced using samples containing overlapping data, these samples cannot be treated as independent. However, any differences in the data should explain differences in the way that aggregation (and the confounding missingness) affects AUROC scores.  We use a Wilcoxon signed-rank test \cite{wilcoxon1945individual} to test for significance between models.

\paragraph{Tasks}
We test on two common baseline clinical machine learning tasks: mortality prediction and length-of-stay (LOS) predicted, realized as a classification task by splitting patients between high LOS ($\ge 3$ days) and low LOS ($< 3$ days).

\paragraph{Experiments}
\label{subsec:training_regimes}
In addition to measuring baseline, year-agnostic performance, we train models for both tasks under 3 regimes, in all cases profiling the performance difference between using the raw \texttt{Item-ID} representation vs. our \texttt{Clinically Aggregated} representation.

\begin{enumerate}[nosep]
    \item \textbf{Year-Specific one-time training}: Train models on the data from 2001 and 2002. If a sudden \texttt{itemid} shift occurs in recording a vital, this model will not recover. 
    \item \textbf{Year-Specific continuous training}: Train a model on \textbf{all} previous years e.g., data from 2001-2005 will be used to train a model that is tested on data from 2006.
    \item \textbf{Year-Specific short-term training}: Train models on the data from \textbf{only} the previous year, e.g., data from 2005 will be used to train a model that is tested on data from 2006.
\end{enumerate}

In addition to these temporal drift experiences, we assess how much performance over time changes as we reduce the size of our overall train set.

\section{Results and Discussion}

\subsection{Performance comparison between feature representations for clinical prediction}
\begin{figure}[h!]
\centering
    \begin{subfigure}{0.3\textwidth}
        \centering
        \includegraphics[width=1.0\linewidth]{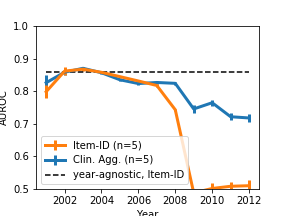}
        \caption{Mortality AUC, models trained on 2001-2002 data.}
        \label{fig:sub1}
    \end{subfigure}\,
    \begin{subfigure}{0.3\textwidth}
        \centering
        \includegraphics[width=1.0\linewidth]{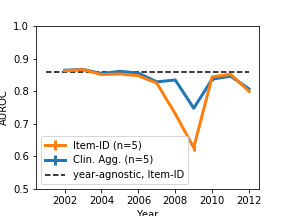}
        \caption{Mortality AUC, models trained yearly on prior year only.}
        \label{fig:sub2}
    \end{subfigure}\,
        \begin{subfigure}{0.3\textwidth}
        \centering
        \includegraphics[width=1.0\linewidth]{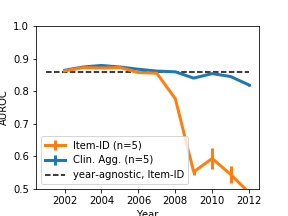}
        \caption{Mortality AUC, models trained yearly on all prior data.}
        \label{fig:sub3}
    \end{subfigure}\\
    \begin{subfigure}{0.3\textwidth}
        \centering
        \includegraphics[width=1.0\linewidth]{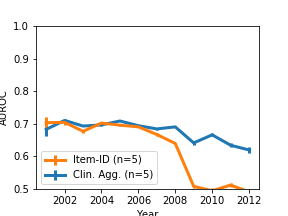}
        \caption{LOS AUC, models trained on 2001-2002 data.}
        \label{fig:los_sub1}
    \end{subfigure}\,
    \begin{subfigure}{0.3\textwidth}
        \centering
        \includegraphics[width=1.0\linewidth]{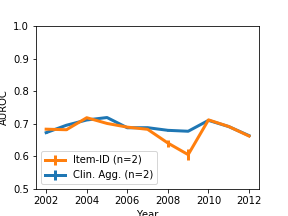}
        \caption{LOS AUC, models trained yearly on prior year only.}
        \label{fig:los_sub2}
    \end{subfigure}\,
    \begin{subfigure}{0.3\textwidth}
        \centering
        \includegraphics[width=1.0\linewidth]{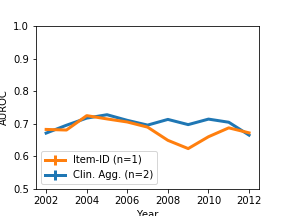}
        \caption{LOS AUC, models trained yearly on all prior data.}
        \label{fig:los_sub3}
    \end{subfigure}\,

    \caption{Performance of RF classifiers using \texttt{Item-Id} and \texttt{Clinically Aggregated} representations on mortality (top) and LOS prediction (bottom). Error bars indicate $\pm$ standard error.}
    \label{fig:itemid_vs_agg}
\end{figure}

We assess mortality prediction and LOS prediction for both of our data representations across all training regimes mentioned in Section~\ref{subsec:training_regimes} in Figure~\ref{fig:itemid_vs_agg}.
%
Overall, we note that the Clinical Aggregate representation is much more robust to the performance degradation over time observed under the \texttt{Item-ID} representation. This resembles the findings in \cite{Gong:2017dh} though our clinically determined groupings appear to offer a lower drop in performance across that shift in practice than do their learned representations.

In Figures \ref{fig:sub1} and \ref{fig:los_sub1} we demonstrate that models trained on the basic \texttt{Item-ID} representation using 2001-2002 data degrade after the hospital system change and do not perform up to the expectations of baseline training paradigms. In Figures \ref{fig:sub2} and \ref{fig:los_sub2}, we further show that training models on the previous year only causes a shock in performance during system change, and Figures \ref{fig:sub3} and \ref{fig:los_sub3} identify that only the \texttt{Clinically Aggregated} representation trained on all prior data is able to sustain the expected performance throughout the system change. 
\subsection{Models Saturate Quickly on Mortality Prediction, Impacting Generalisation}
\begin{figure}[h!]
\centering
    \begin{subfigure}{0.45\textwidth}
        \centering
        \includegraphics[width=1.0\linewidth]{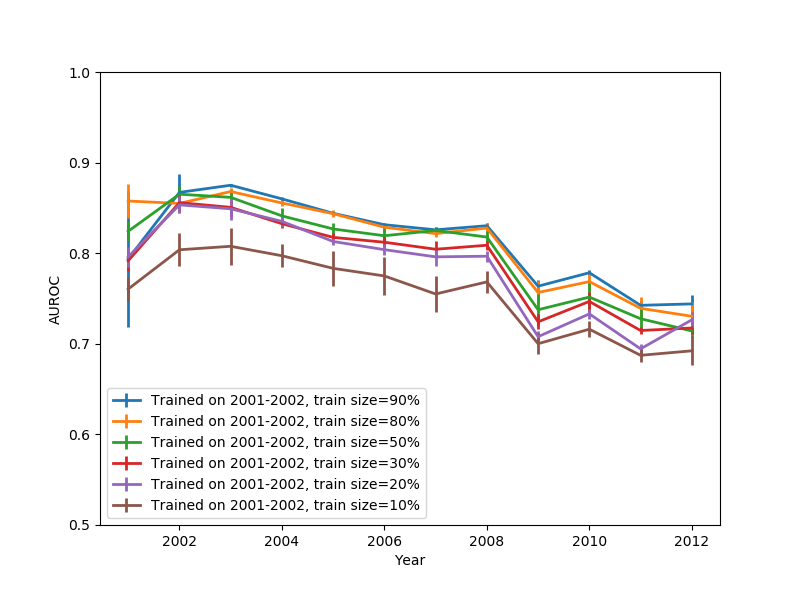}
        \caption{AUROC}
        \label{fig:gen_sub1}
    \end{subfigure}\,
    \begin{subfigure}{0.45\textwidth}
        \centering
        \includegraphics[width=1.0\linewidth]{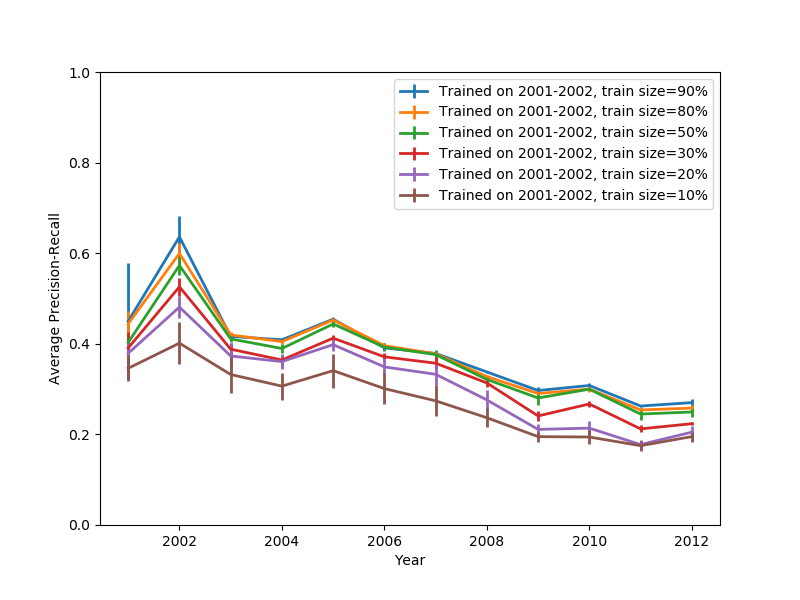}
        \caption{AUPRC}
        \label{fig:gen_sub2}
    \end{subfigure}\,
    \caption{Generalisability of RF models trained on partial data from 2001-2002 alone in ICU mortality prediction. As training data from is reduced from 1,982 to 220 patients (90\% to 10\% training samples), performance decreases by only 7\% for predicting patient outcomes 10 years into the future.}
    \label{fig:saturation}
\end{figure}

We note that predictive performance in RF models trained with \texttt{Clinically Aggregated} representations on only 2001-2002 data performs at a higher level than expected. Specifically, an RF model trained on 90\% of 2001-2002 data (1982 patients) is able to predict mortality ten years later (2012) with an AUROC of 0.744 $\pm$ 0.019 (AUPR 0.270 $\pm$ 0.008). Surprisingly, when the training data was reduced to only 10\% (220 patients) this was observed to drop to AUROC of 0.692 $\pm$ 0.032 (AUPR of 0.195 $\pm$ 0.012) (Figure \ref{fig:saturation}). 

\begin{figure}[h!]
    \includegraphics[width=1.0\textwidth]{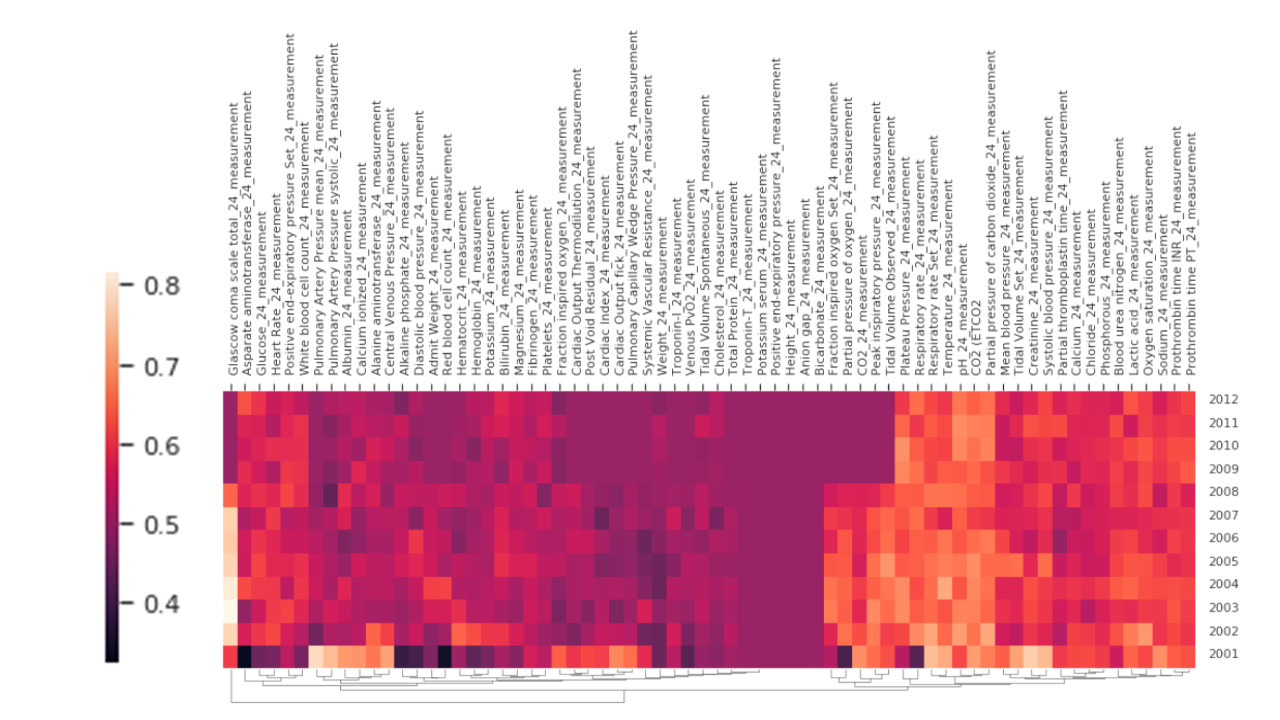}
    \centering
    \caption{Random forest classifiers trained on 2001-2002 data of a single feature measured over 24 hours using the \texttt{Item-ID} representation (n=5 per feature)}
     \label{fig:ablation}
\end{figure}

Such prediction results imply that mortality may be a trivial prediction task. Using feature ablations, we demonstrate that one feature, Glasgow coma scale, alone is able to sustain AUROCs greater than 0.77 for the duration in which it is measured (Figure~\ref{fig:ablation}). 

\section{Conclusion}
While it is essential to compare models on standardised tasks, we should not expect to train models that translate into the clinic on MIMIC features without careful consideration of how data was generated~\cite{ghassemi2018opportunities}. We demonstrate that only using a continuous training regime of models, with a \texttt{Clinically Aggregated} representation can generate yearly performances that are comparable to those reported when training year-agnostic models.
\label{headings}

\printbibliography

\newpage

\begin{appendices}
\section{Feature aggregation description}
\label{app:feat_agg}
    For example, in MIMIC III, \texttt{CHARTEVENTS} with \texttt{itemid}s \textit{861, 1127, 1542, and 220546} are averaged into one feature in the \texttt{Clinically Aggregated} feature vector called \textit{White blood cell count}. Note that \texttt{itemid}s \textit{861, 1127, and 1542} were recorded in the 2001-2008 system and \texttt{itemid} \textit{220546} was recorded in the 2008-2012 system.

\section{Significance test for representation generalisability}
\label{app:sig_generalizability}
\begin{figure}[H]
    \includegraphics[width=0.5\textwidth]{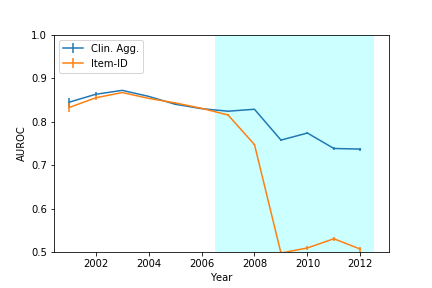}
    \centering
    \caption{Area under the ROC for the task of classifying patient mortality in ICU, after observing 24 hours of data. Shaded regions indicate significant differences between \texttt{Clinically Aggregated} representation and \texttt{Item-ID} representation (Wilcoxon signed-rank test, p<0.01, n=20)}
\end{figure}

\end{appendices}
\end{document}